\documentclass{article}
\usepackage{spconf,amsmath,graphicx,url,amssymb,amsfonts}
\usepackage{xspace}
\usepackage{colortbl}
\usepackage{color}
\usepackage{float}
\usepackage[table,dvipsnames]{xcolor}
\usepackage{multicol,multirow}
\usepackage{arydshln}
\usepackage{booktabs}
\usepackage{tabularx}
\usepackage{subcaption}
\usepackage{caption}
\usepackage{enumitem,kantlipsum}

\makeatletter
\DeclareRobustCommand\onedot{\futurelet\@let@token\@onedot}
\def\@onedot{\ifx\@let@token.\else.\null\fi\xspace}
\def\eg{\emph{e.g}\onedot} 
\def\ie{\emph{i.e}\onedot}

\makeatother

\title{Object-conditioned Bag of Instances for\\Few-Shot Personalized Instance Recognition}

\name{Umberto Michieli{$^{1}$,} Jijoong Moon{$^{2}$,}  Daehyun Kim{$^{2}$,} Mete Ozay{$^{1}$}}
\address{$^{1}$\,Samsung Research UK, $^{2}$\,Samsung Research Korea}
\begin{document}
\maketitle
\begin{abstract}
Nowadays, users demand for increased personalization of vision systems to localize and identify personal \textit{instances} of objects (\eg, \textit{my dog} rather than \textit{dog}) from a few-shot dataset only.
Despite outstanding results of deep networks on classical label-abundant benchmarks (\eg, those of the latest YOLOv8 model for standard object detection), they struggle to maintain within-class variability to represent different instances rather than object categories only.
We construct an \textit{Object-conditioned Bag of Instances} (OBoI) based on multi-order statistics of extracted features, where generic object detection models are extended to search and identify personal instances from the OBoI's metric space, without need for backpropagation. 
By relying on multi-order statistics, OBoI achieves consistent superior accuracy in distinguishing different instances. 
In the results, we achieve $77.1\%$ personal object recognition accuracy in case of $18$ personal instances, showing about $12\%$ relative gain over the state of the art. 
\end{abstract}
\section{Introduction}
\label{sec:intro}

Smart devices are starting to be ubiquitous in everyday life \cite{vermesan2022internet} and their users are demanding for instance-level personalized detection of vision systems mounted on such devices \cite{he2019device,arora2021value}. 
For example, vacuum cleaners can now monitor the behavior of users' specific pets, and stay away from those specific pets that are mostly scared by the robot's noise \cite{vacuum_cleaners}.
Nonetheless, users do not provide many labeled examples, being a time-consuming operation.
Therefore, we introduce a new task of few-shot instance-level personalization of object detection models to detect and recognize personal instances of objects (\eg, \textit{dog}$_1$ and \textit{dog}$_2$ rather than just \textit{dog}).
The limited availability of the data distinguishes our task from previous instance-level personalization attempts \cite{camoriano2017incremental,lomonaco2017core50}.
To the best of our knowledge, previous works assume large availability of labelled data and finetune (FT) the models through computationally expensive updates. %
However, FT-based methods inevitably fail when few-shot samples are provided \cite{wang2017few,jung2022few,zhu2023uncertainty,abdali2023active}.

In our work, we utilize the latest YOLOv8 \cite{yolov8} efficient detection model, and we enable personalized instance recognition via backpropagation-free Prototypes-based Few-Shot Learners (PFSLs), such as \cite{snell2017prototypical,wang2019simpleshot}.
In short, PFSLs  learn  a  metric  space  in  which  classification  is  performed by computing distances to prototypical representations of each class.

In this context, we extend PFSLs to support object-class conditioned search, and we call these approaches \textit{Object-conditioned Bag of Instances} (OBoI), since they contain instance-level prototypes.
Our approach enriches any OBoI method by augmenting localized encoder embeddings (EEs) of the input object via multi-order statistics to construct a richer metric space, where instance-specific patterns are separable. We compute augmented EEs (AEEs) via a reduction module similar to recent pooling schemes \cite{yang2023few,michieli2023online1,michieli2023online2,michieli2024HOP} to characterize the distribution of the specific instances from the few-shot labelled data. 
A concurrent work \cite{yang2023few} applies ensemble learning on multi-order features learned separately; however, their focus is neither personalized instance recognition nor object detection, and they require gradient-based training.
A backpropagation-free approach, instead, could be especially useful where dynamic compilers are not available for the target hardware. 
Our OBoIs with AEEs significantly increase model personalization, alleviating neural collapse \cite{kothapalli2023neural,papyan2020prevalence}, \ie, a state at which within-class variability of hidden layer outputs is completely lost due to the object-level optimization objective. Our main novelties are: 
\begin{enumerate}[noitemsep,nolistsep,leftmargin=*]
    \item We propose a novel task of few-shot personalization of object detectors to recognize instances of objects; 

\item  We extend PFSLs via object-level conditioning (OBoIs); 

\item We further design a multi-order feature space where personal instances can be separated via a backpropagation-free metric learning on few-shot labelled user data only; 

\item OBoIs provide superior results on both same and other domain data (11-22\% and 7-18\% relative gains respectively).
\end{enumerate}

\begin{figure}
    \centering
    \includegraphics[trim=0cm 11cm 15.1cm 0cm, clip, width=\linewidth]{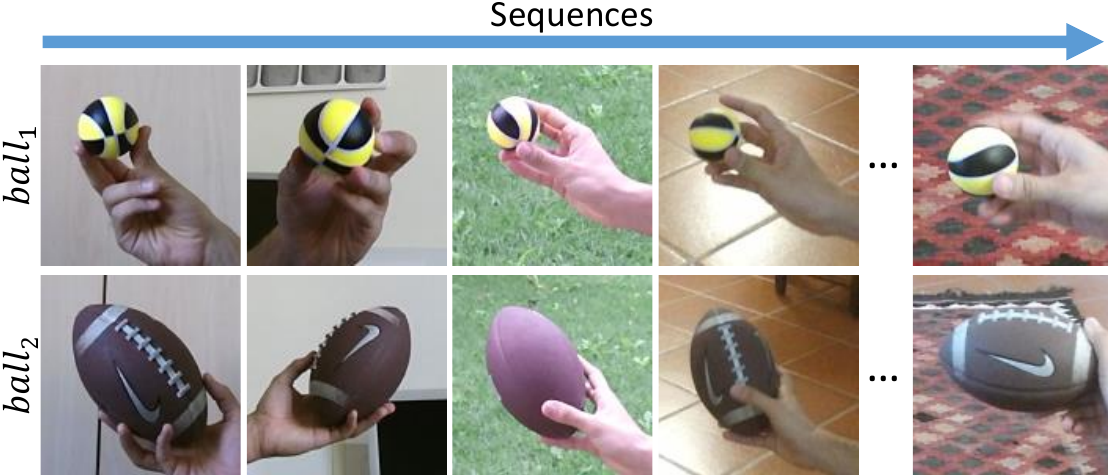}
    \caption{For each object (\eg, \textit{ball}), multiple instances (\eg, \textit{ball}$_1$, \textit{ball}$_2$) are acquired across different sequences.}
    \label{fig:acquisition_sequences}
\end{figure}

\section{Few-Shot Personalized Instance Recognition}
In our setup, we aim at personalizing generic object detection models to recognize objects by a set of instance-level labels. 

We are given a generic object detection model $M_o$ (\eg, YOLOv8 \cite{yolov8}) which has been trained on a labelled object-level dataset $\mathcal{T}_o=\{x_{o,k}, y_{o,k}\}_{k=1}^{N_o}$, whose labels $y_{o,k}\in \mathcal{C}_o$ belong to an object-level class set $\mathcal{C}_o$ (\eg, $\mathcal{C}_o = \{\mathrm{ball}, \mathrm{bottle}\}$). 
We target personalization of the classification ability of $M_o$ to recognize a set of personal classes given few-shot labelled samples $\mathcal{T}_i=\{x_{i,k}, y_{i,k}\}_{k=1}^{N_i}$, where ${N_o \gg N_i}$ and labels $y_{i,k} \in \mathcal{C}_i$ belong to an instance-level class set ${\mathcal{C}_i}$ (\eg, ${\mathcal{C}_i = \{\mathrm{ball}_{\mathrm{1}}, \mathrm{ball}_{\mathrm{2}}, \mathrm{bottle}_{\mathrm{1}} \}}$). We remark that, in this work, we focus on the detector's classification part and we do not update the localization part. In other words, we assume that there exists an \textit{instance-to-object} function $f(\cdot)$ mapping each class $c \in \mathcal{C}_i$ to a label in $\mathcal{C}_o$; \ie, $f: \mathcal{C}_i \to \mathcal{C}_o$ (\eg, $f(\mathrm{ball}_{\mathrm{1}})=\mathrm{ball}$), therefore there is no need to update the detector's regression part.

In the  analyses, we implement the model $M_o$ by YOLOv8, pre-train the model on MSCOCO \cite{lin2014microsoft} and then on $\mathcal{T}_o$, which contains samples from Open-Images-V7 (OIV7) \cite{kuznetsova2020open} of the same object-level classes as in the personal dataset $\mathcal{T}_i$; \ie, such that $f(\cdot)$ exists. We used the default learning parameters \cite{yolov8} for pre-training.
We design several setups whereby we assign a few samples to the training set and the remaining ones to the testing ($80\%$) and validation ($20\%$) sets.

\noindent\textbf{Datasets.} We use CORe50 \cite{lomonaco2017core50} or iCubWorld Transformations (iCWT) \cite{pasquale2019we} as the personalized recognition datasets.
\textit{CORe50:} we consider a subset of 45 personal instances (\ie, $|\mathcal{C}_i|=45$) belonging to 9 object-level classes (\ie, $|\mathcal{C}_o|=9$), acquired over 11 variable-background sequences, \ie, different domains (see Fig.~\ref{fig:acquisition_sequences}).
\textit{iCWT:} we consider a subset of 9 object-level classes with 10 personal instances each acquired under 5 sequences with diverse affine transformations of the items.
On both datasets, we restrict the personalization stage to the frames being correctly labelled by YOLOv8n, maintaining a balanced number of samples per instance and per sequence. 

\noindent\textbf{Metrics.}
We compute the instance recognition accuracy averaged within each object-level class ($\mathrm{Acc}_o$, \%) and averaged over all instances ($\mathrm{Acc}_i$, \%) on the test set. We define the relative gain between two methods obtaining $\mathrm{Acc}_{i,1}$ and $\mathrm{Acc}_{i,2}$ ($\mathrm{Acc}_{i,2}>\mathrm{Acc}_{i,1}$) as: $\Delta \triangleq 100 \cdot (\mathrm{Acc}_{i,2}-\mathrm{Acc}_{i,1}) / (\mathrm{Acc}_{i,1})$.

\begin{table*}[t]
\begin{minipage}{.61\textwidth} %
    \centering
    \includegraphics[trim=0cm 13cm 14.1cm 0cm, clip, width=1\linewidth]{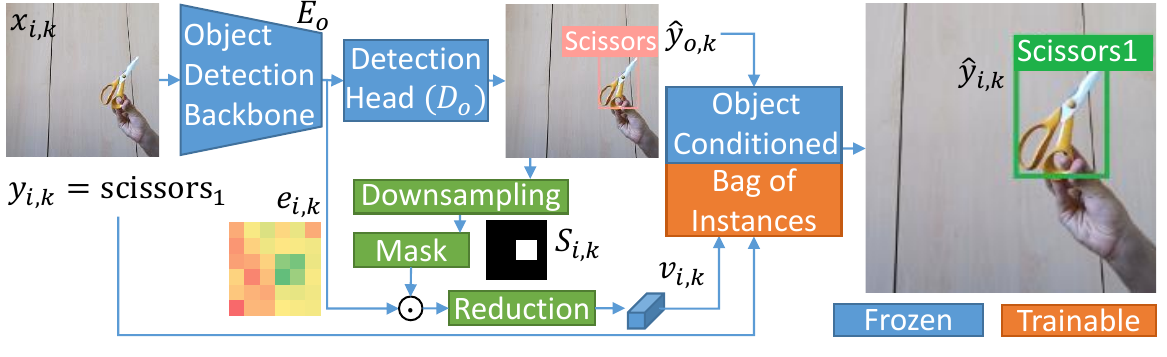}
    \captionof{figure}{%
    A generic object detector (\eg, YOLOv8) is adapted to detect personal instances via our backpropagation-free Object-conditioned Bag of Instances (OBoI) approach with augmented embeddings by multi-order statistics.}
    \label{fig:method}
\end{minipage}\hfill
\begin{minipage}{0.385\textwidth} %
\setlength{\tabcolsep}{4.5pt}
  \centering
  \caption{$\mathrm{Acc}_i$ of OBoIs on 1SAS (same domain).}
  \vspace{-0.2cm}
    \begin{tabular}{lcccc}
    \toprule
    \multirow{2}[0]{*}{} & \multicolumn{4}{c}{\# Instances per object class} \\
          & \textbf{2}     & \textbf{3}     & \textbf{4}    & \textbf{5} \\\midrule %
    Random & 50.00 & 33.33 & 25.00 & 20.00 \\
    FT    & 54.12 & 34.43 & 24.11 & 21.41 \\\hdashline
    SimpleShot & 68.84 & 52.69 & 42.44 & 38.46 \\
    + AEE (ours) & 76.34 & 62.04 & 51.79 & 46.30 \\
    $\Delta$ & \color{OliveGreen} +10.9  & \color{OliveGreen} +17.7  & \color{OliveGreen} +22.0  & \color{OliveGreen} +20.4  \\\hdashline
    ProtoNet & 68.84 & 54.88 & 47.97 & 43.98 \\
    + AEE (ours) & \textbf{77.08} & \textbf{64.57} & \textbf{55.81} & \textbf{53.39} \\
    $\Delta$ & \color{OliveGreen} +12.0  & \color{OliveGreen} +17.7  & \color{OliveGreen} +16.3  & \color{OliveGreen} +21.4  \\
    \bottomrule
    \end{tabular}%
  \label{tab:1shot_allseq_variable_instances_same}%
\end{minipage}
\end{table*}%

\section{Object-conditioned Bag of Instances}

We propose a lightweight module that can be integrated into any object detection network. Our solution is based on three key components: (i) an object detection network, with (ii) a multi-order statistical augmentation of embeddings for (iii) instance-level recognition via an OBoI. Next, we outline how we construct our OBoI to personalize object detectors pre-trained on $\mathcal{T}_o$ on the server side (see Fig.~\ref{fig:method}).

In our case, $\mathcal{T}_o$ is made of subsets of the generic OIV7 dataset. 
The model $M_o$ is then adapted to recognize personal user-specific object instances. 
Without loss of generality, we assume that $M_o=D_o\circ E_o$ can be decomposed into an encoder $E_o$ and a detection head $D_o$. 
Given a sample from $\mathcal{T}_i$, $(x_{i,k}, y_{i,k})\in\mathcal{T}_i$, where ${x_{i,k}\in\mathbb{R}^{H\times W \times 3}}$ is an RGB input image with size $H\times W$ and $y_{i,k}$ the associated personal instance label, we pass $x_{i,k}$ through the model obtaining the object-level predicted label ($\hat{y}_{o,k}$) and bounding box. %

We rescale the predicted coordinates by $H/H'=W/W'$ to match the low-resolution spatial dimensions $H'\times W'$ of the {$D$-dimensional} features ${e_{i,k} \triangleq E_o(x_{i,k}) \in \mathbb{R}^{H'\times W' \times D}}$ (for the sake of simplicity, we assume that only one object is present in each image, but the same rationale applies in presence of multiple objects seamlessly). 
We then build a binary mask $S_{i,k}$ to discard regions outside the bounding box, and we apply it via the Hadamard product obtaining ${e'_{i,k}=S_{i,k} \odot e_{i,k}}$ which is then passed through a reduction operation. In order to characterize the instance-level distribution from the few input samples, we extract and concatenate the first $R$ statistical moments to form $v_{i,k} \triangleq \mathrm{concat}(m_1, \dots, m_R)$ being $\mathrm{concat}(\cdot)$ the concatenation operation and $m_n$ the $n$-th order central statistical moment \cite{papoulis2002probability} computed over the features in $e'_{i,k}$ corresponding to non-zero entries in $S_{i,k}$.

Finally, we use the vector $v_{i,k}$ as the input to some PFSL.
In our case, we use PFSLs to identify instance-level classes on a metric space spanned by multi-order statistics. 
We refer to \cite{snell2017prototypical,wang2019simpleshot} for more details.
Additionally, we condition the search for representations of personal objects only within the instances whose object category matches $\hat{y}_{o,k}$, thus simplifying the search of the correct nearest instance-level prototype.

The overall pipeline can be thought of as an \textit{Object-conditioned Bag of Instances} (OBoI) since generic category-level output is converted to specific personal-level output via conditional nearest prototype selection.
Our setup and method are fully compatible with the key requirement of continually learning new instances over time \cite{lomonaco2017core50,camoriano2017incremental,michieli2023online2,michieli2021continual,michieli2019incremental,lesort2020continual}; whenever a user presents a new instance to be recognized, we can include new instance-level prototypes in the OBoI at any time with no accuracy degradation with respect to the case where all instances are available from the beginning of the adaptation process.

\section{Experimental Results}

Most of the evaluation focuses on the personal instance-level accuracy, since our modules do not influence in any way the object-level detection accuracy and bounding box regression.
For the sake of clarity, we consider that each input sample contains one single instance.
Nonetheless, our method can handle multiple instances in input samples by running the 
instance-level prototype search for each input object independently. Provided that the general object detection results are accurate, our results would not change.

Unless otherwise stated, we report all results on YOLOv8n, being the most suitable for deployed applications, and in the case of 2 instances per each object-level category.

\begin{figure*}[t]
\centering
\begin{subfigure}{.495\linewidth}
  \centering
    \includegraphics[trim=0.2cm 0.2cm 0.2cm 0.2cm, clip, width=\linewidth]{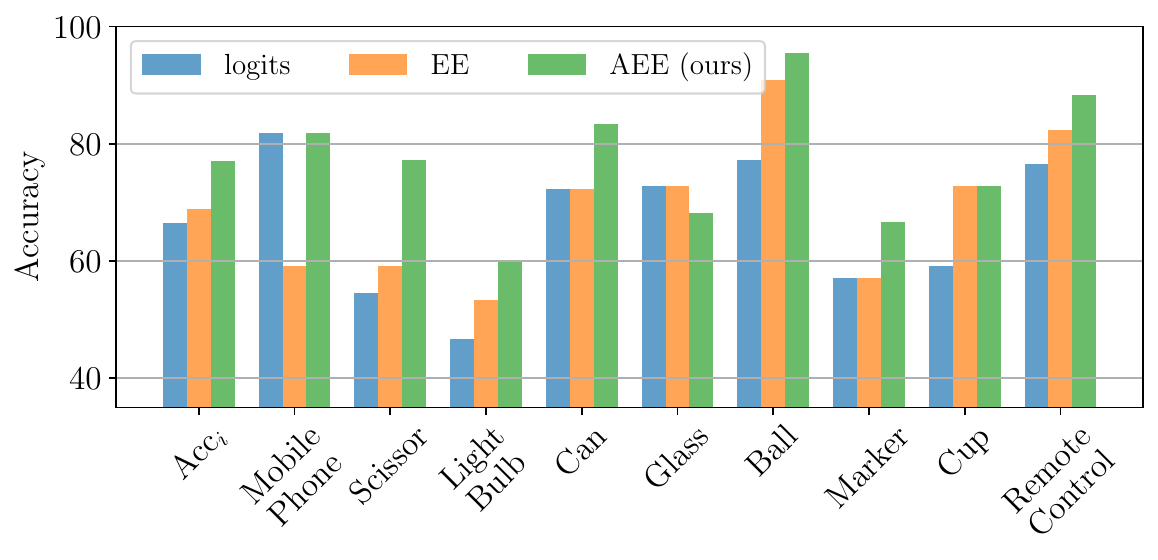}
    \vspace{-0.8cm}
    \caption{1SAS (same domain).}
    \label{fig:per_class_accuracy_same}
\end{subfigure}
\hspace*{\fill}
\begin{subfigure}{.495\linewidth}
\centering
    \includegraphics[trim=0.2cm 0.2cm 0.2cm 0.2cm, clip, width=\linewidth]{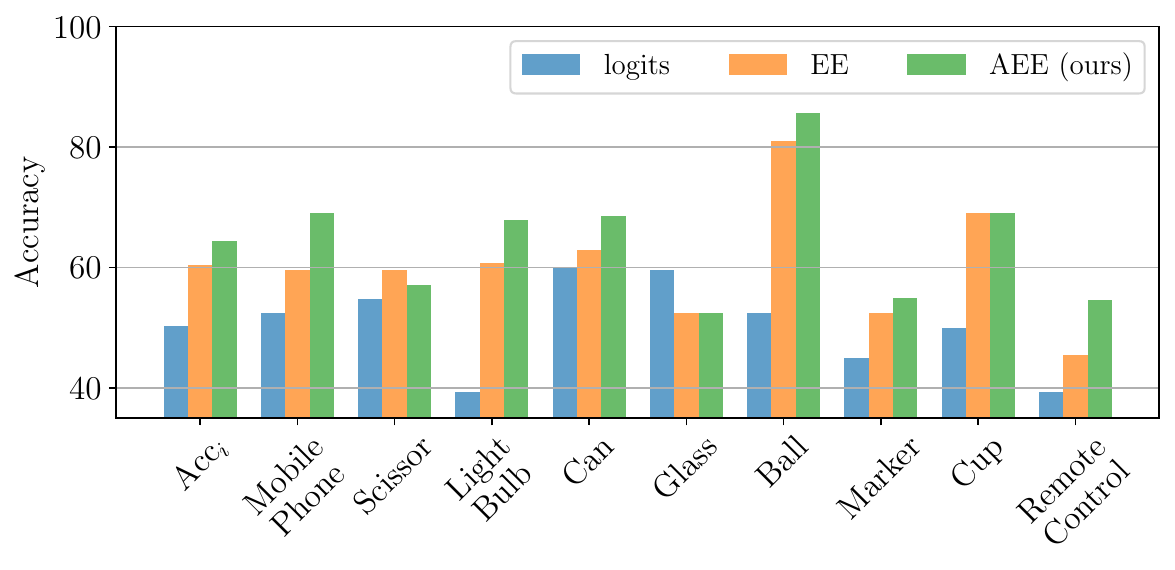}
    \vspace{-0.8cm}
    \caption{1S1S (other domain).}
    \label{fig:per_class_accuracy_other}  
\end{subfigure}
\vspace{-0.4cm}
\caption{$\mathrm{Acc}_i$ and per-object $\mathrm{Acc}_o$ for OBoIs via ProtoNet. EE: encoder embeddings. AEE: augmented EE.}%
\end{figure*}

\textbf{Same domain.} The first scenario we design considers \textbf{1}-\textbf{S}hot from \textbf{A}ll \textbf{S}equences (1SAS), therefore the same domain is seen during few-shot training (one sample per each sequence) and testing (all remaining samples from all sequences). 
Table~\ref{tab:1shot_allseq_variable_instances_same} reports $\mathrm{Acc}_i$ in multiple setups having a variable number of instances per object-level class. First, we observe that gradient-based fine-tuning methods (\eg, FT) are not effective and obtain comparable results to a random classifier (lower bound). OBoI via PFSL methods such as SimpleShot \cite{wang2019simpleshot} and ProtoNet \cite{snell2017prototypical} show large gains compared to FT by learning a metric space from the extracted features.
In both cases, the augmentation of embeddings via our multi-order statistics boost the recognition accuracy significantly, especially in presence of multiple instances per  object. 
Remarkably, we can personalize YOLOv8n to achieve $77.08\%$ $\mathrm{Acc}_i$ when detecting $18$ personal instances via just a few samples and via a backpropagation-free approach, assuming that a correct object-level classification and bounding box regression are output from the detection head.
Fig.~\ref{fig:per_class_accuracy_same} reports $\mathrm{Acc}_i$ and $\mathrm{Acc}_o$ of OBoIs via ProtoNet in the case of 2 instances per object. We consider three configurations for ProtoNet: at the logits level (\ie, the output of the last layer of the detector's head), at the encoder embedding level (\ie, the output of the detector's encoder) or via our multi-statistics augmented encoder embeddings (\ie, with features augmented via multi-order statistics). We observe that our proposed solution consistently improves or achieves comparable results on every object class.

\begin{table}[!t]
\setlength{\tabcolsep}{8pt}
  \centering
  \caption{$\mathrm{Acc}_i$ of OBoIs on 1S1S (other domain).}
  \vspace{-0.15cm}
    \begin{tabular}{lcccc}
    \toprule
    \multirow{2}[0]{*}{} & \multicolumn{4}{c}{\# Instances per object class} \\
          & \textbf{2} & \textbf{3} & \textbf{4} & \textbf{5} \\\midrule
    Random & 50.00 & 33.33 & 25.00 & 20.00 \\
    FT    & 48.12 & 35.98 & 21.74 & 22.53 \\\hdashline
    SimpleShot & 60.62 & 43.95 & 34.31 & 29.84 \\
    + AEE (ours) & \textbf{65.46} & 49.62 & 40.52 & 35.33 \\
    $\Delta$ & \color{OliveGreen} +8.0   & \color{OliveGreen} +12.9  & \color{OliveGreen} +18.1  & \color{OliveGreen} +18.4  \\\hdashline
    ProtoNet & 60.33 & 46.52 & 37.13 & 32.13 \\
    + AEE (ours) & 64.37 & \textbf{51.39} & \textbf{40.70} & \textbf{36.77} \\
    $\Delta$ & \color{OliveGreen} +6.7   & \color{OliveGreen} +10.5  & \color{OliveGreen} +9.6   & \color{OliveGreen} +14.4  \\
    \bottomrule
    \end{tabular}%
  \label{tab:1shot_1seq_variable_instances_other}%
  \vspace{-0.1cm}
\end{table}%

\textbf{Other domain.} We designed a more realistic, yet challenging, setup considering \textbf{1}-\textbf{S}hot from \textbf{1} \textbf{S}equence (1S1S) during training and all remaining samples during testing. The model experiences different domains at training time (one sample from the first sequence only) and at testing time (all remaining samples from all sequences).
Table~\ref{tab:1shot_1seq_variable_instances_other} summarizes the main results. Reducing the training samples further decreased the accuracy of FT, compared to the 1SAS setup. Also SimpleShot and ProtoNet show lower accuracy, due to the fewer training samples and the domain gap of the 1S1S setup. Nonetheless, they exhibit large gains over FT. Our method shows a significant improvement of the performance in every case, even in the presence of a domain shift, and especially in case of multiple instances per object category.
We argue that the gain attained by our approach is lower than the previous one due to the difficulty in reliably matching multi-order statistics between a single input sample from a single domain and several target samples from several domains.
Fig.~\ref{fig:per_class_accuracy_other} reports $\mathrm{Acc}_o$; similarly to the previous case, we confirm that our solution obtains robust results across most of the classes.

\begin{figure}[t]
    \centering
    \includegraphics[trim=0.2cm 0.2cm 0.2cm 0.2cm, clip, width=\linewidth]{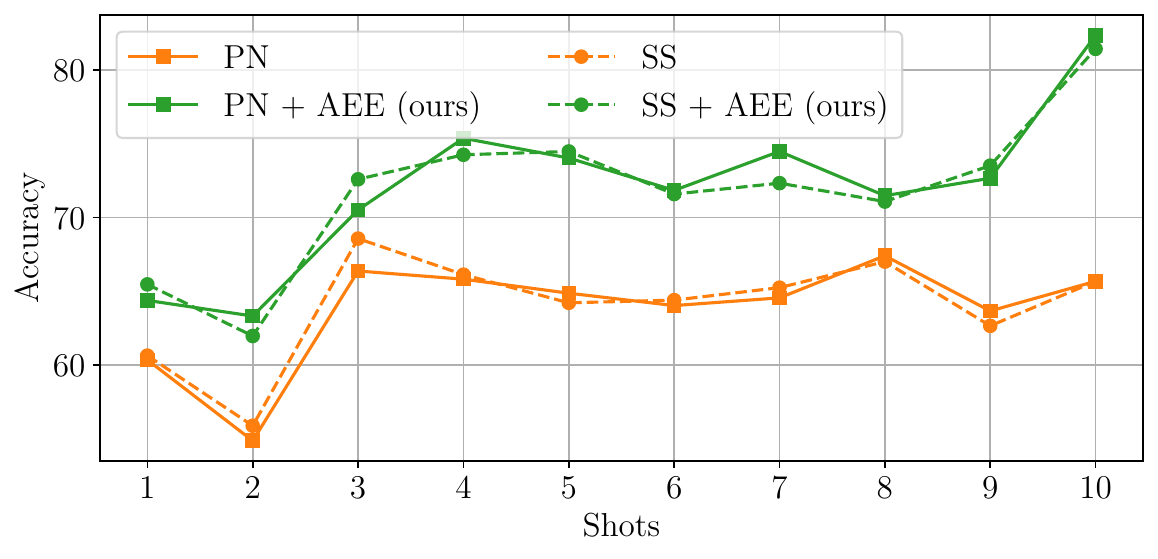}
    \caption{$\mathrm{Acc}_i$ at variable shots. Samples are drawn randomly from each sequence. PN: ProtoNet, SS: SimpleShot.}
    \label{fig:per_shot_accuracy}
\end{figure}

\textbf{Variable training shots} are studied in Fig.~\ref{fig:per_shot_accuracy}, where we observe that OBoIs with our AEE improve personal recognition accuracy regardless of the number of available training samples (\ie, shots) for both ProtoNet and SimpleShot.

\begin{table}[t]
  \centering
  \setlength{\tabcolsep}{2.7pt}
  \caption{Ablation on YOLOv8 models. 
  General object detection results are computed on the subset from OIV7.
  Personal instance recognition is evaluated on the subset from CORe50; we report a single score that is the $\mathrm{mean}_p(\mathrm{Acc}_{i,p})$ where $p$ indicates the setup with $p$ instances per object ($p\in\{2,3,4,5\}$).}
    \begin{tabular}{clccccc}
    \toprule
        \multicolumn{2}{c}{YOLOv8:}      & \textbf{n}     & \textbf{s}     & \textbf{m}     & \textbf{l}     & \textbf{x} \\\midrule
          & Size [MB] & 5.9   & 21.4  & 77.0 & 83.6  & 130.4 \\\hline
    \multirow{4}[0]{*}{\rotatebox[origin=c]{90}{\textbf{Detection}}} & Precision & 72.0  & 73.2  & 70.4  & 72.4  & 77.2 \\
          & Recall & 72.5  & 70.3  & 74.4  & 75.5  & 77.3 \\
          & mAP50 & 77.7  & 77.6  & 79.3  & 78.2  & 78.7 \\
          & mAP50-95 & 60.9  & 61.7  & 62.3  & 62.7  & 63.1 \\\midrule
    \multirow{3}[0]{*}{\rotatebox[origin=c]{90}{\textbf{1S1S}}} & ProtoNet & 44.03 & 43.56 & 48.44 & 55.38 & 58.77 \\
          & + AEE (ours) & \textbf{48.31} & \textbf{48.47} & \textbf{56.30} & \textbf{56.88} & \textbf{60.84} \\
          & $\Delta$ & \color{OliveGreen} +9.72  & \color{OliveGreen} +11.27 & \color{OliveGreen} +16.23 & \color{OliveGreen} +2.72  & \color{OliveGreen} +3.52  \\\hdashline
    \multirow{3}[0]{*}{\rotatebox[origin=c]{90}{\textbf{1SAS}}} & ProtoNet & 53.92 & 53.86 & 61.66 & 64.16 & 66.60 \\
          & + AEE (ours) & \textbf{62.71} & \textbf{66.67} & \textbf{75.29} & \textbf{75.21} & \textbf{74.65} \\
          & $\Delta$ & \color{OliveGreen} +16.31 & \color{OliveGreen} +23.79 & \color{OliveGreen} +22.10 & \color{OliveGreen} +17.23 & \color{OliveGreen} +12.09  \\\bottomrule
    \end{tabular}%
    \vspace{-0.3cm}
  \label{tab:yolos}%
\end{table}%

\textbf{Other YOLOv8 sizes} are evaluated in Table~\ref{tab:yolos} on both general object detection and personalized instance recognition task.
Larger YOLOv8 models can improve detection performance, and this correlates with the personal instance recognition accuracy. The improvement of larger YOLOv8
\begin{table}[H]
  \centering
  \setlength{\tabcolsep}{1.6pt}
  \caption{$\mathrm{Acc}_i$ on iCWT. PN: ProtoNet.}
    \begin{tabular}{clccccccccc}
    \toprule
    \multicolumn{2}{c}{\multirow{2}[0]{*}{}} & \multicolumn{9}{c}{\textbf{\# Instances per object class}} \\
    \multicolumn{2}{c}{} & \textbf{2} & \textbf{3} & \textbf{4} & \textbf{5} & \textbf{6} & \textbf{7} & \textbf{8} & \textbf{9} & \textbf{10} \\
    \midrule
    \multirow{3}[0]{*}{\rotatebox[origin=c]{90}{1SAS}} & PN & 82.3  & 66.5  & 62.6  & 58.8  & 46.9  & 46.0  & 43.6  & 41.5  & 41.5 \\
          & + AEE  & 85.8  & 75.8  & 71.9  & 67.2  & 57.7  & 55.8  & 52.4  & 49.5  & 47.8 \\
          & $\Delta$ & \color{OliveGreen}\footnotesize+4.4   & \color{OliveGreen}\footnotesize+13.9  & \color{OliveGreen}\footnotesize+14.9  & \color{OliveGreen}\footnotesize+14.3  & \color{OliveGreen}\footnotesize+22.9  & \color{OliveGreen}\footnotesize+21.5  & \color{OliveGreen}\footnotesize+20.2  & \color{OliveGreen}\footnotesize+19.1  & \color{OliveGreen}\footnotesize+15.2 \\\midrule
    \multirow{3}[0]{*}{\rotatebox[origin=c]{90}{1S1S}} & PN & 71.7  & 52.9  & 51.7  & 47.2  & 39.5  & 38.7  & 36.1  & 33.4  & 31.1 \\
          & + AEE  & 79.6  & 60.1  & 56.5  & 49.8  & 41.1  & 40.6  & 37.5  & 34.2  & 33.0 \\
          & $\Delta$ & \color{OliveGreen}\footnotesize+11.1  & \color{OliveGreen}\footnotesize+13.5  & \color{OliveGreen}\footnotesize+9.4   & \color{OliveGreen}\footnotesize+5.5   & \color{OliveGreen}\footnotesize+4.1   & \color{OliveGreen}\footnotesize+5.0   & \color{OliveGreen}\footnotesize+3.8   & \color{OliveGreen}\footnotesize+2.4   & \color{OliveGreen}\footnotesize+6.1 \\\bottomrule
    \end{tabular}%
  \label{tab:iCWT}%
\end{table}%
\noindent models comes at a cost of a significantly larger model size and slower FPS: YOLOv8x improves personal recognition by about $25\%$ compared to YOLOv8n, while having about 22$\times$ larger size and $3.6\times$ slower inference. The final choice depends on the  hardware specifications of target devices.

\textbf{Computational inference time} of our AEE on top of the OBoI with ProtoNet increases by as little as $0.8\%$ making our method lightweight with a nearly negligible impact. %

\textbf{Additional ablation} studies to evaluate our design choices are reported here on ProtoNet. Removing the object-level conditioning lowered $\mathrm{Acc}_i$ in the 1SAS setup from $77.1\%$ of our approach (ProtoNet + AEE) to $70.9\%$, showing a relative gain of about $3\%$ compared to the baseline ($68.8\%$). This is due to the larger search space in metric learning. Removing the mask $S_{i,k}$ leads to more background noise regions to flow into prototype computation, and it decreases accuracy by $2.6\%$ in the 1SAS setup, and by even more ($5.1\%$) in the 1S1S setup, since background varies across different sequences.

\textbf{Another personal dataset (iCWT)} is shown in Table~\ref{tab:iCWT} against the highest baseline ProtoNet. Our approach exhibits robust gains across all setups ranging from 18 to 90 instances.

\section{Conclusion}
In this paper, we introduced few-shot instance-level personalization of object detectors. 
We proposed a new method (OBoI) to personalize detection models to recognize user specific instances of object categories.
OBoI is a backpropagation-free metric learning approach on a multi-order statistics feature space.
We believe that this setup and our method could pave the way to personal instance-level detection and could stimulate future research and applications.

\vfill\pagebreak

\bibliographystyle{IEEEbib}
\bibliography{strings_short,umbib}

\end{document}